\pdfoutput=1

\documentclass[11pt]{article}

\usepackage{ACL2023}

\usepackage{times}
\usepackage{latexsym}
\usepackage{multirow}
\usepackage{paralist}
\usepackage[T1]{fontenc}

\usepackage[utf8]{inputenc}

\usepackage{microtype}

\usepackage{inconsolata}

\usepackage{booktabs} 
\usepackage{siunitx} 
\usepackage{tcolorbox} 

\title{\textsc{LLM-Eval}: Unified Multi-Dimensional Automatic Evaluation for Open-Domain Conversations with Large Language Models}

\author{
  Yen-Ting Lin, Yun-Nung Chen \\
  National Taiwan University, Taipei, Taiwan \\
  \texttt{\{ytl, y.v.chen\}@ieee.org} \\
}

\begin{document}
\maketitle
\begin{abstract}
We propose \textsc{LLM-Eval}, a unified multi-dimensional automatic evaluation method for open-domain conversations with large language models (LLMs). Existing evaluation methods often rely on human annotations, ground-truth responses, or multiple LLM prompts, which can be expensive and time-consuming. To address these issues, we design a single prompt-based evaluation method that leverages a unified evaluation schema to cover multiple dimensions of conversation quality in a single model call. We extensively evaluate the performance of \textsc{LLM-Eval} on various benchmark datasets, demonstrating its effectiveness, efficiency, and adaptability compared to state-of-the-art evaluation methods. Our analysis also highlights the importance of choosing suitable LLMs and decoding strategies for accurate evaluation results. \textsc{LLM-Eval} offers a versatile and robust solution for evaluating open-domain conversation systems, streamlining the evaluation process and providing consistent performance across diverse scenarios.
\end{abstract}

\begin{figure}[h!]
    \centering
    \includegraphics[width=1.0\linewidth]{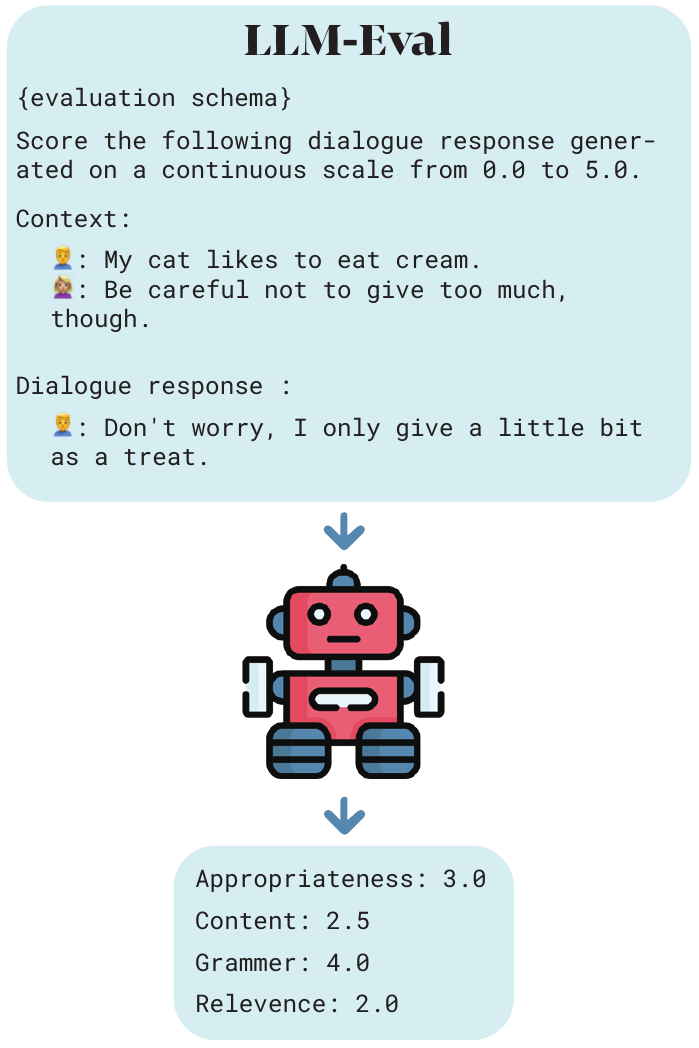}
    \vspace{-10pt}
    \caption{An illustration of our proposed \textsc{LLM-Eval} framework, which leverages a unified multi-dimensional evaluation schema and a single prompt to efficiently evaluate open-domain conversations with large language models.}
    \vspace{-17pt}
    \label{fig:what}
\end{figure}

\section{Introduction}

Effective evaluation of open-domain conversation systems is a critical yet challenging problem in natural language processing research \cite{smith-etal-2022-human}. Accurate and consistent evaluation methods are essential for understanding and improving the performance of dialogue systems. Traditional automatic evaluation metrics, such as BLEU \cite{papineni-etal-2002-bleu} and ROUGE \cite{lin-2004-rouge}, are insufficient for capturing the nuances of natural language conversations \cite{liu-etal-2016-evaluate,DBLP:journals/air/DeriuROERAC21}, leading to the development of various advanced metrics \cite{DBLP:conf/aaai/TaoMZY18,ghazarian-etal-2019-better,sai-etal-2020-improving, huang-etal-2020-grade, mehri-eskenazi-2020-usr, phy-etal-2020-deconstruct, zhang-etal-2021-dynaeval, li-etal-2021-conversations, DBLP:journals/corr/abs-2302-04166,DBLP:journals/corr/abs-2303-16634}. However, most existing methods require annotation data , human references, or multiple prompts, which could be expensive, time-consuming, or prone to errors.

In this paper, we address the problem of evaluating open-domain conversation systems with a focus on large language models (LLMs) (Figure \ref{fig:what}). Our goal is to develop an efficient and accurate evaluation method that covers multiple dimensions of conversation quality, such as content, grammar, relevance, and appropriateness, without requiring human references or multiple prompts. We build upon recent advances in LLMs \cite{DBLP:conf/nips/BrownMRSKDNSSAA20,DBLP:journals/corr/abs-2204-05862,DBLP:journals/corr/abs-2303-08774}, and propose a unified multi-dimensional evaluation method called \textsc{LLM-Eval}.

Existing evaluation methods have demonstrated promising results in various aspects of dialogue evaluation. However, they often rely on human annotations \cite{mehri-eskenazi-2020-usr,phy-etal-2020-deconstruct}, ground-truth responses \cite{DBLP:conf/aaai/GhazarianWGP20, DBLP:conf/iwsds/ZhangDBFL20}, or multiple LLM inferences \cite{DBLP:journals/corr/abs-2302-04166,DBLP:journals/corr/abs-2303-16634}, limiting their efficiency and adaptability in practical scenarios. We aim to bridge this gap by proposing \textsc{LLM-Eval}, a single-prompt-based evaluation method that leverages a unified evaluation schema to cover multiple dimensions of conversation quality in a single model call.

In \textsc{LLM-Eval}, we design a natural language instruction that defines the evaluation task and desired criteria, as well as a format instruction that specifies the structure and range of scores for each dimension. The single prompt is created by concatenating the dialogue context, reference (if available), and generated response, and then fed to a large language model, which outputs scores for each dimension based on the defined schema. 

We extensively evaluate the performance of \textsc{LLM-Eval} on a variety of benchmark datasets, covering diverse dialogue systems and evaluation dimensions. Our experiments demonstrate that \textsc{LLM-Eval} consistently outperforms most baselines and state-of-the-art evaluation methods in terms of correlation with human judgments. The proposed method is also robust and versatile, adapting to different scoring ranges and evaluation scenarios.

In summary, our main contributions are as follows:
\begin{itemize}
\item We propose \textsc{LLM-Eval}, a unified multi-dimensional automatic evaluation method for open-domain conversations with large language models, which streamlines the evaluation process by using a single prompt and a unified evaluation schema.
\item We extensively evaluate the performance of \textsc{LLM-Eval} on a variety of benchmark datasets, demonstrating its effectiveness and efficiency in comparison with state-of-the-art evaluation methods.
\item We provide an in-depth analysis of the impact of different LLMs and decoding methods on the performance of \textsc{LLM-Eval}, highlighting the importance of choosing suitable LLMs and decoding strategies for accurate evaluation results.
\end{itemize}

\section{Related Work}

\paragraph{Multi-Dimensional Metrics}
Multi-dimensional evaluation metrics have been proposed to assess various aspects of dialogue quality, such as content, grammar, relevance, and appropriateness. Examples include USR \cite{mehri-eskenazi-2020-usr}, which trains multiple models to measure qualities like fluency, relevance, and knowledge conditioning, and GRADE \cite{huang-etal-2020-grade}, which models topic transition dynamics in dialogue history using a graph representation. FlowScore \cite{li-etal-2021-conversations} leverages dynamic information flow in dialog history to measure dialogue quality. Unlike these approaches, \textsc{LLM-Eval} employs a single prompt-based evaluation method that leverages a unified evaluation schema, streamlining the evaluation process and providing a more efficient and adaptable solution.

\paragraph{Unsupervised Metrics}
Unsupervised evaluation metrics aim to assess the quality of dialogue responses without requiring human annotations. Notable unsupervised methods include DEB \cite{sai-etal-2020-improving}, which fine-tunes BERT with an NSP objective on a dataset with relevant and adversarial irrelevant responses, and FED \cite{mehri-eskenazi-2020-unsupervised}, an unsupervised method that measures dialogue quality using features derived from response embeddings and language model probabilities. In contrast, \textsc{LLM-Eval} leverages the power of large language models to provide a unified multi-dimensional evaluation, achieving better performance and adaptability compared to existing unsupervised methods.

\paragraph{Large Language Models for Evaluation}
Recent works have explored using large language models for dialogue evaluation. GPTScore \cite{DBLP:journals/corr/abs-2302-04166} employs models like GPT-3 to assign higher probabilities to quality content, using multiple prompts for a multi-dimensional assessment. 
\citet{DBLP:journals/corr/abs-2304-00723} explores using ChatGPT and InstructGPT to evaluate text quality without references, and compares different paradigms of using LLMs, including generating explicit scores, using model confidence to determine implicit scores, and directly comparing pairs of texts.
G-EVAL \cite{DBLP:journals/corr/abs-2303-16634}, a framework that leverages LLMs with chain-of-thoughts (CoT)\cite{DBLP:conf/nips/Wei0SBIXCLZ22} and a form-filling paradigm. G-EVAL with GPT-4 as the backbone model achieves a high correlation with human judgments on a summarization task. However, both GPTScore and G-EVAL require multiple prompts or complex scoring functions that use probabilities of output tokens and their weighted summation as the final score, which can be inefficient or time-consuming. \textsc{LLM-Eval} addresses these issues by using a single prompt and a unified evaluation schema, offering a more efficient and adaptable evaluation method for open-domain conversations. Additionally, \textsc{LLM-Eval} provides multi-dimensional evaluation scores in a single model call, further streamlining the evaluation process.

\section{Methodology}

\textsc{LLM-Eval} is an efficient prompt-based evaluator tailored for open-domain conversations with large language models. It encompasses a single prompt that addresses the evaluation task, desired evaluation criteria, and a unified multi-dimensional evaluation schema. This method eradicates the necessity for numerous LLMs inferences or intricate scoring functions \cite{DBLP:journals/corr/abs-2302-04166,DBLP:journals/corr/abs-2303-16634}, while still delivering a comprehensive assessment of the generated text.

\paragraph{Unified Evaluation Schema} The evaluation schema is a natural language instruction that defines the task and the desired evaluation criteria. It is designed to cover multiple dimensions of the evaluation, such as content, grammar, relevance, and appropriateness. The schema is provided as a format instruction, which specifies the structure and the range of the scores for each dimension. For example, the evaluation schema can be:

\begin{quote}
\emph{Human: The output should be formatted as a JSON instance that conforms to the JSON schema below.
...
Here is the output schema:
\{"properties": \{"content": \{"title": "Content", "description": "content score in the range of 0 to 100", "type": "integer"}, "grammar": ...\}
\end{quote}

\paragraph{Single Prompt for Evaluation} The single prompt is designed to include the necessary dialogue context and the target response that needs to be evaluated, along with the evaluation schema. The prompt is concatenated with the dialogue context, the reference (if available), and the generated response, and then fed to the large language model to output a score for each evaluation dimension, based on the defined schema. For example, the prompt for evaluating a dialogue response with human reference can be:

\begin{quote}
\emph{Context: \{context\} \\Reference: \{reference\} \\Dialogue response: \{response\}}
\end{quote}

\paragraph{Efficient Evaluation} By using a single prompt with a unified evaluation schema, \textsc{LLM-Eval} can efficiently obtain multi-dimensional scores for the responses without the need for multiple prompts. The large language model is called only once, and it directly provides the evaluation scores for each dimension based on the defined schema. For instance, given a dialogue context, reference, and generated response, the \textsc{LLM-Eval} method would produce an example output that looks like this:

\begin{quote}
\emph{Output: \{"appropriateness": 3.0, "content": 2.5, "grammar": 4.0, "relevance": 2.0\}}
\end{quote}

This output showcases the multi-dimensional evaluation of the generated response, with each dimension receiving a score based on the predefined schema. The scores help in understanding the quality of the response in terms of appropriateness, content, grammar, and relevance, while still maintaining the efficiency of the evaluation process by requiring just a single call to the large language model.
For a detailed description of the prompt templates used in our experiments with \textsc{LLM-Eval}, please refer to Appendix \ref{sec:appendix_prompt}.

\section{Experiments}

\subsection{Datasets and Benchmarks}

Our proposed \textsc{LLM-Eval} method is assessed on an array of datasets spanning diverse dialogue systems and evaluation dimensions. We provide a concise overview of the datasets and their features in this section. The datasets include human annotations, where each entry comprises a dialogue context, a generated response, and associated scores. A ground-truth human reference may also be present. For data lacking human reference, we only evaluate reference-free metrics.

\paragraph{DSTC10 Hidden Set} The DSTC10 hidden set \cite{DBLP:journals/corr/abs-2111-02110} is a multi-dimensional evaluation dataset that includes JSALT \cite{DBLP:conf/iwsds/Kong-VegaSWD18}, NCM, ESL \cite{DBLP:journals/corr/VinyalsL15, sedoc-etal-2019-chateval,DBLP:journals/corr/abs-2010-12741}, Topical-DSTC10 \cite{DBLP:conf/interspeech/GopalakrishnanH19} and Persona-DSTC10 \cite{zhang-etal-2018-personalizing}. JSALT contains human-generated dialogue segments from EmpatheticDialogues \cite{rashkin-etal-2019-towards} and TopicalChat \cite{DBLP:conf/interspeech/GopalakrishnanH19}. NCM and ESL are datasets with pairwise comparisons between system responses, collected from an English learning website and hand-crafted prompts. Topical-DSTC10 and Persona-DSTC10 are newly created datasets that include responses from various dialogue systems, such as LSTM Seq2Seq, HRED, VHRED, BlenderBot, DialoGPT, T5, and GPT-3.

\paragraph{Overall Scores with Human Reference} TopicalChat-USR evaluates response quality in knowledge-grounded dialogues, emphasizing topical understanding. PersonaChat-USR measures response quality in personalized conversations, highlighting the incorporation of speaker personas \cite{mehri-eskenazi-2020-usr}. ConvAI2-GRADE examines the quality of chit-chat dialogue systems, focusing on engaging and contextually relevant responses. DailyDialog-GRADE investigates response quality in everyday conversational contexts. EmpatheticDialogue-GRADE assesses the quality of empathetic responses in dialogue systems \cite{huang-etal-2020-grade}. DSTC6 evaluates end-to-end conversation modeling with human-generated responses \cite{DBLP:journals/corr/HoriH17}.

\paragraph{Overall Scores without Human Reference} DailyDialog-PredictiveEngagement evaluates engagement in dialogue systems without relying on human references \cite{DBLP:conf/aaai/GhazarianWGP20}. FED is an unsupervised method that measures the quality of dialogue responses without using human references \cite{mehri-eskenazi-2020-unsupervised}. DSTC9 focuses on the end-to-end evaluation of context-aware dialogue systems without human references \cite{mehri-etal-2022-interactive}.

We compare the performance of \textsc{LLM-Eval} with existing evaluation methods on these datasets to demonstrate its effectiveness and efficiency in evaluating open-domain conversations. The evaluation results are presented in terms of correlation with human judgments, using Pearson's correlation coefficient ($r$) and Spearman's correlation coefficient ($\rho$).

\subsection{\textsc{LLM-Eval} Configurations}
We evaluate \textsc{LLM-Eval} under different settings to demonstrate its effectiveness and adaptability. The configurations are as follows:
\paragraph{\textsc{LLM-Eval 0-5}} The evaluation scores for each dimension are in the range of 0 to 5 with one decimal place, which is more close to common 1-5 Likert scale used in human evaluation.
\paragraph{\textsc{LLM-Eval 0-100}} The evaluation scores for each dimension are in the range of 0 to 100 as integers, providing a finer-grained scale for evaluation.

The evaluation schema prompt for both configurations remains the same, with only the range of scores differing between them. We test the \textsc{LLM-Eval} method with and without human references for each configuration if applicable.

Unless specified otherwise, throughout our experiments and evaluations, we employ the Anthropic Claude API with the \texttt{claude-v1.3} model and use greedy decoding, which selects the token with the highest probability at each time step during the generation process.

\begin{table*}[h!]
\small
\centering
\begin{tabular}{lccc|cccc|cccc|c}
\toprule
\multirow{2}{*}{\bf Spearman $\rho$ (\%)} & \textbf{JSALT} & \textbf{ESL}   & \textbf{NCM}   & \multicolumn{4}{c|}{\textbf{TopicalChat-DSTC10}} & \multicolumn{4}{c|}{\textbf{PersonaChat-DSTC10}} & \multirow{2}{*}{\textbf{Avg} }\\
& \textbf{APP}   & \textbf{APP}   & \textbf{APP}   & \textbf{APP} & \textbf{CON} & \textbf{GRA} & \textbf{REL}   & \textbf{APP}               & \textbf{CON}       & \textbf{GRA}                 & \textbf{REL}                           &       \\
\midrule
Deep-AM-FM & ~~5.1 & 32.3 & 16.5 & 18.2& ~~9.4 & 17.9& 26.2  & 21.0 & 14.7 & 19.1 & 24.1 & 18.4\\
DSTC10 Team 1 & \textbf{27.7} & 42.0 & 29.9 & 29.7 & ~~7.0 & 11.6 & 37.0 & 38.6 & 19.3 & 18.6 & 44.5 & 30.2 \\
MME-CRS & 11.7&41.4 & 29.9 & 32.6 & 17.2 & ~~9.0 & \textbf{44.8} & 45.6 & 32.5 & 22.0 & \textbf{54.8}& 31.0\\
\midrule
\multicolumn{5}{l}{\textit{\textbf{without human reference}}} \\
\textsc{LLM-Eval} {\tiny \textit{0-5}}  & 23.2 & 51.8  & \textbf{34.4} & \bf 38.6 & 20.6 & \textbf{33.2} &\underline{42.8}& \bf 48.2 & \underline{36.9}& \bf 34.5 & \underline{52.1} & \bf 37.8 \\
\textsc{LLM-Eval} {\tiny \textit{0-100}}        & \underline{27.3} & 50.5 & \underline{34.2} & \textbf{38.6} & 21.3 & \underline{32.7} & 41.1 & 47.6&\textbf{37.8}& 30.2 & 51.9 & \underline{37.6}\\
\multicolumn{5}{l}{\textit{\textbf{with human reference}}} \\
\textsc{LLM-Eval} {\tiny \textit{0-5}} & 25.4  & \underline{51.8}   & 32.5  & 38.0 & \underline{21.5}   & 31.2   & 42.2     &\underline{47.9}& 36.0   & \underline{30.6}   & 49.1 & 36.9 \\
\textsc{LLM-Eval} {\tiny \textit{0-100}} & 25.7& \textbf{51.9} & 30.8 & \underline{38.2} & \textbf{21.6}   & 30.0   & 40.2     & 45.4  & 34.8   & 28.6   & 49.3 & 36.0 \\
\bottomrule
\end{tabular}
\caption{Spearman correlation coefficients between human ratings and automatic metrics across multiple dimensions (\textit{APP} for Appropriateness, \textit{CON} for Content, \textit{GRA} for Grammar, and \textit{REL} for Relevance) for DSTC10 hidden test datasets with human reference. Each team is represented by the best submission on 5 test datasets. The best score for each column is highlighted in bold. The second best is underlined. Note that the last column is averaged over 11 dimension-wise correlation scores of all five datasets.}
\label{tab:results_hidden}
\end{table*}

\begin{table*}[t!]
  \small
  \centering
  \begin{tabular}{lcccccc|c}
    \toprule
    \bf $r$ / $\rho$ (\%) & \textbf{TopicalChat} & \textbf{PersonaChat} & \textbf{ConvAI2} & \textbf{DD} & \textbf{ED} & \textbf{DSTC6} & \textbf{Average}\\
    \midrule
    BLEU-4 & 21.6 / 29.6  & 13.5  / ~~9.0  & ~~0.3 / 12.8 & ~~7.5  / 18.4 & ~-5.1 / ~~0.2& 13.1  / 29.8 & ~~8.5 / 16.6\\
ROUGE-L & 27.5 / 28.7  & ~~6.6  / ~~3.8  & 13.6 / 14.0 & 15.4 / 14.7 & ~~2.9 / ~-1.3 & 33.2 / 32.6 & 16.5 / 15.4\\
BERTScore        & 29.8  / 32.5  & 15.2  / 12.2  & 22.5 / 22.4 & 12.9 / 10.0 & ~~4.6 / ~~3.3  & 36.9  / 33.7 & 20.3 / 19.0\\
DEB             & 18.0  / 11.6  & 29.1  / 37.3  & 42.6 / 50.4 & \underline{33.7} / \textbf{36.3}  & 35.6 / 39.5  & 21.1  / 21.4 & 30.0 / 32.8\\
GRADE            & 20.0  / 21.7  & 35.8  / 35.2  & 56.6 / 57.1 & 27.8 / 25.3 & 33.0 / 29.7 & 11.9 / 12.2 & 30.9 / 30.2\\
USR          & 41.2  / 42.3  & 44.0  / 41.8  & 50.1 / 50.0 & ~~5.7 / ~~5.7 & 26.4 / 25.5 & 18.4  / 16.6 &  31.0 / 30.3\\
USL-H            & 32.2  / 34.0  & 49.5  / 52.3  & 44.3 / 45.7 & 10.8 / ~~9.3 & 29.3 / 23.5 & 21.7  / 17.9 & 31.3 / 30.5\\
    \midrule
    \multicolumn{5}{l}{\textit{\textbf{without human reference}}} \\
    \textsc{LLM-Eval} {\tiny \textit{0-5}} & \underline{55.7} / \underline{58.3} & 51.0 / 48.0 & \underline{59.3} / \underline{59.6} & 31.8 / 32.2 & 42.1 / 41.4 & 43.3 / 41.1 & \underline{47.2} / 46.8\\
    \textsc{LLM-Eval} {\tiny \textit{0-100}} & 49.0 / 49.9 & 53.3 / 51.5 & \textbf{61.3} / \textbf{61.8} & \textbf{34.6} / \underline{34.9} & \underline{43.2} / \underline{42.3} & 44.0 / 41.8 & \textbf{47.6} / \underline{47.0}\\
    \multicolumn{5}{l}{\textit{\textbf{with human reference}}} \\
    \textsc{LLM-Eval} {\tiny \textit{0-5}} & \textbf{56.5} / \textbf{59.4} & \textbf{55.4} / \textbf{53.1} & 43.1 / 43.8 & 32.0 / 32.2 & 40.0 / 40.1 & \underline{47.0} / \underline{45.5} & 45.7 / 45.7\\
    \textsc{LLM-Eval} {\tiny \textit{0-100}} & 55.6 / 57.1 & \underline{53.8} / \underline{52.7} & 45.6 / 45.9 & 33.4 / 34.0 & \textbf{43.5} / \textbf{43.2} & \textbf{49.8} / \textbf{49.9} & 47.0 / \bf 47.1\\
    \bottomrule
  \end{tabular}
  \caption{Correlation coefficients (Pearson $r$ and Spearman $\rho$) between human ratings and automatic metrics in terms of overall scores for datasets with human reference. We use the following abbreviations: TopicalChat (TopicalChat-USR), PersonaChat (PersonaChat-USR), ConvAI2 (ConvAI2-GRADE), DD (DailyDialog-GRADE), ED (EmpatheticDialogue-GRADE). The best score for each column is highlighted in bold. The second best is underlined.}
  \label{tab:results_ref}
\end{table*}

\subsection{Baseline Evaluation Metrics}

We compare \textsc{LLM-Eval} with several state-of-the-art evaluation metrics, including both traditional and LLM-based approaches.

\begin{compactitem}
    \item \textbf{Deep-AM-FM} measures dialog quality with Adequacy Metric (AM) and Fluency Metric (FM), utilizing BERT embeddings and language model probabilities \cite{DBLP:conf/iwsds/ZhangDBFL20}.
    \item \textbf{DSTC10 Team 1} boosted DyanEval's \cite{zhang-etal-2021-dynaeval} turn-level evaluation performance by integrating auxiliary objectives and combining USL-H\cite{phy-etal-2020-deconstruct}, DEB \cite{sai-etal-2020-improving}, and an improved DyanEval, with weights based on input dialogue data characteristics \cite{DBLP:journals/corr/abs-2111-02110}.
    \item {\bf MME-CRS} introduces the Multi-Metric Evaluation, consisting of 5 parallel sub-metrics to assess dialogue quality across fluency, relevance, engagement, specificity, and topic coherence. The approach utilizes Correlation Re-Scaling to model sub-metric relationships \cite{DBLP:journals/corr/abs-2206-09403}.
    \item {\bf BERTScore} computes the F1 score by matching token embeddings in human references and system responses \cite{DBLP:conf/iclr/ZhangKWWA20}.
    \item {\bf DEB} constructs a dialog dataset with relevant and adversarial irrelevant responses, then fine-tunes BERT with an NSP objective \cite{sai-etal-2020-improving}.
    \item {\bf GRADE} models topic transition dynamics in dialog using a graph representation of the dialog history \cite{huang-etal-2020-grade}.
    \item {\bf USR} trains several models to measure different qualities of dialogs, including fluency, relevance, and knowledge conditioning \cite{mehri-eskenazi-2020-usr}.
    \item {\bf USL-H} combines three models trained with different objectives (VUP, NSP, MLM) to evaluate response validity, sensibleness, and likelihood \cite{phy-etal-2020-deconstruct}.
    \item {\bf DynaEval} leverages a graph structure to model dialog-level interactions between user and system \cite{zhang-etal-2021-dynaeval}.
    \item {\bf FlowScore} models dynamic information flow in dialog history and measures dialog quality using DialoFlow representations \cite{li-etal-2021-conversations}.
    \item {\bf GPTScore} evaluates text using models like GPT-3, assigning higher probabilities to quality content through multiple prompts for a multi-dimensional assessment. However, it may not be as effective as \textsc{LLM-Eval}, which only requires a single prompt \cite{DBLP:journals/corr/abs-2302-04166}.
    \item {\bf Traditional Metrics}: We also include classic metrics such as BLEU \cite{papineni-etal-2002-bleu} and ROUGE \cite{lin-2004-rouge}, which have known limitations in dialogue evaluation.
\end{compactitem}

\begin{table*}[t]
\centering
\begin{tabular}{lcccc|c}
\toprule
\multirow{2}{*}{\textbf{$r$ / $\rho$ (\%)}}   & \textbf{DailyDialog-PE} &  \multicolumn{2}{c}{\textbf{FED}} & \textbf{DSTC9} & \multirow{2}{*}{\textbf{Average}}\\
& \textbf{Turn-Level} &\textbf{Turn-Level} &\textbf{Dialog-Level} & \textbf{Dialog-Level} \\
\midrule
DynaEval & 16.7 / 16.0 & 31.9 / 32.3 & 50.3 / 54.7 & ~~9.3 / 10.1 & 27.1 / 28.3\\
USL-H & 68.8 / 69.9 & 20.1 / 18.9 & ~~7.3 / 15.2 & 10.5 / 10.5 & 26.7 / 28.6\\
FlowScore & - & ~-6.5 / ~-5.5 & ~-7.3 / ~-0.3 & 14.7 / 14.0 & ~~0.3 / ~~2.7\\
\midrule
GPTScore & - & ~~~-~~~~/ 38.3 & ~~~-~~~ / 54.3 & -& ~~~-~~~ / 46.3\\
\textsc{LLM-Eval} {\tiny \textit{0-5}} & \underline{71.0} / \textbf{71.3} & \textbf{60.4} / \textbf{50.9} & \textbf{67.6} / \textbf{71.4} & \underline{15.9} / \underline{16.5} & \textbf{53.7} / \textbf{52.5}\\
\textsc{LLM-Eval} {\tiny \textit{0-100}} & \textbf{71.4} / \underline{71.0} & \underline{59.7} / \underline{49.9} & \underline{64.4} / \underline{70.4} & \textbf{16.1} / \textbf{18.6} & \underline{52.9} / \underline{52.5}\\
\bottomrule
\end{tabular}
\caption{Correlation coefficients (Pearson $r$ and Spearman $\rho$) between human ratings and automatic metrics in terms of overall scores for datasets without human reference. The best score for each column is highlighted in bold. The second best is underlined.}
\label{tab:results_noref}
\end{table*}

\subsection{Results of DSTC10 Hidden Set}

The results of our proposed \textsc{LLM-Eval} method on the DSTC10 hidden set are presented in Table \ref{tab:results_hidden}. We compare the performance of \textsc{LLM-Eval} with other participating teams and baselines in the DSTC10 challenge. The evaluation is performed in terms of Spearman correlation coefficients between human ratings and automatic metrics across multiple dimensions, including Appropriateness (APP), Content (CON), Grammar (GRA), and Relevance (REL).

The results show that \textsc{LLM-Eval} consistently outperforms most of the baselines and even the best performing team in DSTC10 across different dimensions and datasets. In particular, \textsc{LLM-Eval} with a 0-5 score range achieves the highest average Spearman correlation coefficient of $0.378$ among all the methods without human reference.

When comparing the two \textsc{LLM-Eval} configurations, both 0-5 and 0-100 settings demonstrate competitive performance, with the 0-5 configuration slightly outperforming the 0-100 configuration in both cases with or without human reference.
This indicates that the \textsc{LLM-Eval} method is robust and versatile in evaluating open-domain conversations, as it can adapt to different scoring ranges and consistently outperform all baselines and the best performing team in DSTC10 across various dimensions and datasets.

\subsection{Overall Scores with Human Reference}
The results of \textsc{LLM-Eval} on datasets with overall scores and human references are presented in Table \ref{tab:results_ref}. We compare the performance of \textsc{LLM-Eval} with other top-performing evaluation methods \cite{yeh-etal-2021-comprehensive}, such as BLEU, ROUGE, BERTScore, DEB, GRADE, USR, and USL-H. The meta-evaluation is performed in terms of Pearson correlation coefficient ($r$) and Spearman correlation coefficient ($\rho$) between human ratings and automatic metrics.

For the DailyDialog-GRADE, ConvAI2-GRADE, and EmpatheticDialogue-GRADE datasets, we use the \textit{"Relevance"} dimension for evaluation, while for the DSTC6 dataset, we use the \textit{``Overall''} score. For TopicalChat-USR and PersonaChat-USR, we predict all the \textit{"Engaging, Maintains Context, Natural, Overall, Understandable, Uses Knowledge"} dimensions in the original annotations but only use the \textit{"Overall"} score for meta-evaluation.

\textsc{LLM-Eval} consistently outperforms most of the baselines across the datasets and correlation coefficients, with LLM-Eval 0-100 configuration achieving the highest average correlation coefficient across all datasets. 

The consistent performance of both configurations across different datasets and dimensions indicates that \textsc{LLM-Eval} is a reliable and effective evaluation tool for open-domain conversations with human references. Its ability to adapt to different scoring ranges while maintaining competitive performance against state-of-the-art evaluation methods showcases the versatility and robustness of the \textsc{LLM-Eval} approach.

\begin{table*}[t]
\small
\centering
\begin{tabular}{lcccc|cccc|c}
\toprule
\multirow{2}{*}{\bf Spearman $\rho$ (\%)} & \multicolumn{4}{c|}{\textbf{Topical-DSTC10}} & \multicolumn{4}{c|}{\textbf{Persona-DSTC10}} & \multirow{2}{*}{\bf Average}\\
& \textbf{APP} & \textbf{CON} & \textbf{GRA} & \textbf{REL}& \textbf{APP}& \textbf{CON}& \textbf{GRA} & \textbf{REL} \\
\midrule
Deep-AM-FM & 18.2& ~~9.4 & 17.9 & 26.2  & 21.0 & 14.7 & 19.1 & 24.1  & 18.9\\
DSTC10 Team 1 & 29.7 & ~~7.0 & 11.6 & 37.0 & 38.6 & 19.3 & 18.6 & 44.5 & 25.8\\
MME-CRS & 32.6 & 17.2 & ~~9.0 & \textbf{44.8} & 45.6 & 32.5 & 22.0 & \textbf{54.8}  & 32.3\\
\midrule
\multicolumn{5}{l}{\textit{\textbf{without human reference}}} \\
 \textsc{LLM-Eval} {\tiny \textit{0-5}} \\
 ~~~\texttt{Anthropic Claude} & \bf 38.6 & 20.6 & \underline{33.2} &\underline{42.8}& \bf 48.2 & \underline{36.9}& \bf 34.5 & \underline{52.1} & \textbf{38.4}\\
~~~\texttt{Anthropic Claude $top\_p=0.9$} & 31.9 & 16.9 & 30.2  & 38.5 & 39.4 & 30.2 & 28.9 & 46.3   & 32.8   \\
~~~\texttt{OpenAI ChatGPT} & 35.7 & 18.4& 33.1 & 37.3& 43.5& 33.4& 30.1 & 48.8  & 35.0\\
~~~\texttt{OpenAI GPT-3.5} & 29.3 & 16.9 & 20.9& 37.1& 36.5& 30.2   & 21.7& 45.2  & 29.7\\
 \textsc{LLM-Eval} {\tiny \textit{0-100}} \\
~~~\texttt{Anthropic Claude} & \textbf{38.6} & 21.3 & 32.7 & 41.1 & 47.6&\textbf{37.8}& 30.2 & 51.9  & \underline{37.7} \\
~~~\texttt{Anthropic Claude $top\_p=0.9$} & 30.1  & 15.6   & 27.3   & 37.7     & 36.2 & 27.9   & 25.9   & 45.4    & 30.8  \\
~~~\texttt{OpenAI ChatGPT} & 36.2& 16.7 & \textbf{33.4}& 36.0 & 44.0& 31.7& 31.4 & 48.1  & 34.7 \\
~~~\texttt{OpenAI GPT-3.5} & 28.2 & 13.9 & 23.5 & 34.0 & 34.8 & 24.7& 21.7 & 42.9 & 28.0\\
\multicolumn{5}{l}{\textit{\textbf{with human reference}}} \\
 \textsc{LLM-Eval} {\tiny \textit{0-5}}\\
 ~~~\texttt{Anthropic Claude} & 38.0 & \underline{21.5}   & 31.2   & 42.2     &\underline{47.9}& 36.0   & 30.6   & 49.1   & 37.1   \\
~~~\texttt{Anthropic Claude-instant} & 26.5 & 14.3& 30.1   & 27.0& 33.4& 30.5   & 25.8   & 35.2    & 27.9  \\
~~~\texttt{OpenAI ChatGPT} & 34.0 & 18.9 & 30.3& 35.1  & 39.4& 30.0& 25.6 & 40.9 & 31.8 \\
~~~\texttt{OpenAI GPT-3.5} & 30.0 & 17.3 & 21.2 & 38.8 & 37.9 & 28.8& 20.8& 45.1  & 30.0 \\
\textsc{LLM-Eval} {\tiny \textit{0-100}}\\
~~~\texttt{Anthropic Claude} & \underline{38.2} & \textbf{21.6}   & 30.0   & 40.2     & 45.4  & 34.8   & 28.6   & 49.3   & 36.0   \\
~~~\texttt{Anthropic Claude-instant} & 28.0 & 14.3 & 32.1   & 34.0     & 37.5     & 31.1   & \underline{32.0}   & 40.8    & 31.2  \\
~~~\texttt{OpenAI ChatGPT}  & 34.6& 20.6 & 31.1 & 35.4 & 39.7& 31.3 & 23.8& 44.1   & 32.6 \\
~~~\texttt{OpenAI GPT-3.5} & 12.4 & 20.8 & 30.5  & 37.8 & 26.6& 20.7 & 24.0 & 40.0  & 26.6\\
\bottomrule
\end{tabular}
\caption{Spearman correlation coefficients between human ratings and \textsc{LLM-Eval} with different configurations across multiple dimensions (\textit{APP} for Appropriateness, \textit{CON} for Content, \textit{GRA} for Grammar, and \textit{REL} for Relevance) for Topical-DSTC10 and Persona-DSTC10. The best score for each column is highlighted in bold. The second best is underlined.}
\label{tab:llm_compa}
\end{table*}

\subsection{Overall Scores without Human Reference}
Table \ref{tab:results_noref} presents the performance of \textsc{LLM-Eval} on datasets without human references, comparing it with other high-performing evaluation methods such as DynaEval, USL-H, and FlowScore.

For the evaluation of DailyDialog-PredictiveEngagement and DSTC9 datasets, we utilize the \textit{"Overall"} score. In the FED dataset, we predict \textit{"Correctness, Engagement, Fluency, Interestingness, Overall, Relevance, Semantically Appropriateness, Specificity, and Understandability"} dimensions for turn-based evaluation, and \textit{"Coherence, Consistency, Topic Depth, Diversity, Error Recovery, Flexibility, Informativeness, Inquisitiveness, Likability, Overall, and Understandability"} dimensions for dialogue-based evaluation. Nonetheless, only the "Overall" score is used for meta-evaluation in each scenario.

Both \textsc{LLM-Eval} configurations, 0-5 and 0-100, consistently display strong performance across the datasets, highlighting their resilience and flexibility. The method's capacity to accommodate different scoring ranges while maintaining competitiveness against state-of-the-art evaluation techniques demonstrates \textsc{LLM-Eval}'s adaptability and robustness. This establishes its value as an efficient and versatile evaluation solution in reference-free settings.

\section{Analysis}

\subsection{Different LLMs}
In this section, we analyze the performance of \textsc{LLM-Eval} when using different large language models for evaluation. Table \ref{tab:llm_compa} presents the Spearman correlation coefficients between human ratings and \textsc{LLM-Eval} with various model configurations and scoring ranges for the Topical-DSTC10 and Persona-DSTC10 datasets. We compare the performance of \textsc{LLM-Eval} when using different LLMs, such as \texttt{Anthropic Claude}, \texttt{OpenAI ChatGPT}, \texttt{Anthropic Claude-instant}, and \texttt{OpenAI GPT-3.5} \footnote{Anthropic Claude (claude-v1.3), OpenAI ChatGPT (gpt-3.5-turbo-0301), Anthropic Claude-instant (claude-instantv1.0), and OpenAI GPT-3.5 (text-davinci-003).}.

Among these models, \texttt{Claude} and \texttt{ChatGPT} are optimized for chat applications, while \texttt{GPT-3.5} is not. We observe that both \texttt{Claude} and \texttt{ChatGPT} generally achieve better performance across all dimensions when compared to \texttt{GPT-3.5}. This suggests that using dialogue-optimized LLMs in the \textsc{LLM-Eval} method leads to more accurate evaluation results in the context of open-domain conversations.

Moreover, when comparing the \texttt{Claude} and \texttt{ChatGPT} models, both models demonstrate competitive performance across different evaluation dimensions, with \texttt{Claude} slightly outperforming \texttt{ChatGPT} in certain configurations.

We also analyze the performance of \texttt{Claude-instant}, a smaller version of \texttt{Claude}. Although it is not as competitive as its larger counterpart, it still achieves reasonable performance in some cases. This implies that smaller models, while not optimal, can still be employed for \textsc{LLM-Eval} to a certain extent, possibly providing a more resource-efficient option in specific scenarios.

In conclusion, our analysis demonstrates that dialogue-optimized LLMs, such as \texttt{Claude} and \texttt{ChatGPT}, yield better performance in the \textsc{LLM-Eval} method for open-domain conversation evaluation. Although smaller models like \texttt{Anthropic Claude-instant} may not achieve the best performance, they can still be considered for resource-limited scenarios. Overall, the choice of LLMs in \textsc{LLM-Eval} plays a crucial role in obtaining accurate evaluation results.

\subsection{Decoding Methods}
In our experiments, we employ greedy decoding for generating responses using the Anthropic API with the \texttt{claude-v1.3} model. Greedy decoding selects the token with the highest probability at each time step during the generation process. However, other decoding methods, such as nucleus sampling could be employed in the \textsc{LLM-Eval} method to explore their impact on the evaluation results.

Nucleus sampling, also known as top-p sampling, samples tokens from the top-$p$ most probable tokens at each time step, where $p$ is a pre-defined probability threshold. This method introduces some randomness into the generation process and could lead to more diverse and creative responses.

Comparing the performance of \texttt{Claude} and \texttt{Claude $top\_p=0.9$} in Table \ref{tab:llm_compa}, we observe that greedy decoding generally achieves better performance across all evaluation dimensions. This finding suggests that using greedy decoding with the \textsc{LLM-Eval} method provides more accurate and consistent evaluation results compared to nucleus sampling. 

One possible reason for this difference in performance is that greedy decoding tends to generate more coherent and focused responses due to its deterministic nature. In contrast, nucleus sampling introduces randomness into the generation process, which may result in less focused or less relevant responses, affecting the evaluation scores. Consequently, greedy decoding appears to be a more suitable choice for the \textsc{LLM-Eval} method.

\section{Conclusion}
In this paper, we introduced \textsc{LLM-Eval}, a unified multi-dimensional automatic evaluation method for open-domain conversations with large language models. The proposed method employs a single prompt along with a unified evaluation schema that covers multiple dimensions of evaluation, such as content, grammar, relevance, and appropriateness. This approach streamlines the evaluation process and eliminates the need for multiple prompts. Experiments on various datasets demonstrated the effectiveness and efficiency of \textsc{LLM-Eval}, consistently outperforming most baselines and state-of-the-art evaluation methods.

As future work, we plan to explore reinforcement learning from LLMs feedback and investigate LLM-in-the-loop evaluation strategies as an alternative to human-in-the-loop methods. This will further enhance the applicability and performance of the \textsc{LLM-Eval} method in various dialogue system evaluation scenarios.

\section*{Limitations}
Although \textsc{LLM-Eval} has shown promising results in assessing open-domain conversations, it is crucial to acknowledge its limitations.

Firstly, the performance of our method relies heavily on the large language models underlying it, which may exhibit biases or generate unexpected outputs. If the language model misinterprets the evaluation schema or prompt instructions, it could lead to inaccurate evaluation scores.

Secondly, the choice of LLM significantly influences the evaluation results, as demonstrated in our analysis. While dialogue-optimized LLMs produce better performance, this selection may limit \textsc{LLM-Eval}'s applicability for particular tasks or dialogue systems.

Thirdly, our approach employs single-number scoring for each evaluation dimension, which may fail to capture the subtleties of human judgments, particularly for subjective aspects like engagement, creativity, or humor.

Lastly, the effectiveness of \textsc{LLM-Eval} hinges on the quality and clarity of the prompts and evaluation schemas. Creating such prompts and schemas may require domain expertise and knowledge of LLM behavior, posing challenges for non-experts.

To overcome these limitations, future research can focus on exploring alternative prompt designs, refining evaluation schemas, and expanding the method to cover a wider range of evaluation dimensions and dialogue system types.

\section*{Ethics Statement}
We acknowledge that there are potential ethical concerns associated with the use of large language models in our evaluation method.

A primary concern is the biases present in large language models. These biases are introduced during the training process, as the models learn from textual data that may contain biased information, stereotypes, or misinformation. When using these biased models for evaluation, it is possible that the evaluation scores produced by \textsc{LLM-Eval} may reflect and perpetuate these biases, potentially leading to biased evaluations of dialogue system outputs. This could, in turn, affect the development of future dialogue systems by encouraging biased behavior.

To mitigate this concern, researchers and developers should be cautious when interpreting the evaluation results obtained through \textsc{LLM-Eval} and consider potential biases in the large language models used. Moreover, future work could explore techniques to debias language models or employ alternative evaluation schemas that actively account for biases in the evaluation process.

\section*{Acknowledgements}

\bibliography{anthology,custom}

\begin{thebibliography}{36}
\expandafter\ifx\csname natexlab\endcsname\relax\def\natexlab#1{#1}\fi

\bibitem[{Bai et~al.(2022)Bai, Jones, Ndousse, Askell, Chen, DasSarma, Drain,
  Fort, Ganguli, Henighan, Joseph, Kadavath, Kernion, Conerly, Showk, Elhage,
  Hatfield{-}Dodds, Hernandez, Hume, Johnston, Kravec, Lovitt, Nanda, Olsson,
  Amodei, Brown, Clark, McCandlish, Olah, Mann, and
  Kaplan}]{DBLP:journals/corr/abs-2204-05862}
Yuntao Bai, Andy Jones, Kamal Ndousse, Amanda Askell, Anna Chen, Nova DasSarma,
  Dawn Drain, Stanislav Fort, Deep Ganguli, Tom Henighan, Nicholas Joseph,
  Saurav Kadavath, Jackson Kernion, Tom Conerly, Sheer~El Showk, Nelson Elhage,
  Zac Hatfield{-}Dodds, Danny Hernandez, Tristan Hume, Scott Johnston, Shauna
  Kravec, Liane Lovitt, Neel Nanda, Catherine Olsson, Dario Amodei, Tom~B.
  Brown, Jack Clark, Sam McCandlish, Chris Olah, Benjamin Mann, and Jared
  Kaplan. 2022.
\newblock \href {https://doi.org/10.48550/arXiv.2204.05862} {Training a helpful
  and harmless assistant with reinforcement learning from human feedback}.
\newblock \emph{CoRR}, abs/2204.05862.

\bibitem[{Brown et~al.(2020)Brown, Mann, Ryder, Subbiah, Kaplan, Dhariwal,
  Neelakantan, Shyam, Sastry, Askell, Agarwal, Herbert{-}Voss, Krueger,
  Henighan, Child, Ramesh, Ziegler, Wu, Winter, Hesse, Chen, Sigler, Litwin,
  Gray, Chess, Clark, Berner, McCandlish, Radford, Sutskever, and
  Amodei}]{DBLP:conf/nips/BrownMRSKDNSSAA20}
Tom~B. Brown, Benjamin Mann, Nick Ryder, Melanie Subbiah, Jared Kaplan,
  Prafulla Dhariwal, Arvind Neelakantan, Pranav Shyam, Girish Sastry, Amanda
  Askell, Sandhini Agarwal, Ariel Herbert{-}Voss, Gretchen Krueger, Tom
  Henighan, Rewon Child, Aditya Ramesh, Daniel~M. Ziegler, Jeffrey Wu, Clemens
  Winter, Christopher Hesse, Mark Chen, Eric Sigler, Mateusz Litwin, Scott
  Gray, Benjamin Chess, Jack Clark, Christopher Berner, Sam McCandlish, Alec
  Radford, Ilya Sutskever, and Dario Amodei. 2020.
\newblock \href
  {https://proceedings.neurips.cc/paper/2020/hash/1457c0d6bfcb4967418bfb8ac142f64a-Abstract.html}
  {Language models are few-shot learners}.
\newblock In \emph{Advances in Neural Information Processing Systems 33: Annual
  Conference on Neural Information Processing Systems 2020, NeurIPS 2020,
  December 6-12, 2020, virtual}.

\bibitem[{Chen et~al.(2023)Chen, Wang, Jiang, Shi, and
  Xu}]{DBLP:journals/corr/abs-2304-00723}
Yi~Chen, Rui Wang, Haiyun Jiang, Shuming Shi, and Ruifeng Xu. 2023.
\newblock \href {https://doi.org/10.48550/arXiv.2304.00723} {Exploring the use
  of large language models for reference-free text quality evaluation: {A}
  preliminary empirical study}.
\newblock \emph{CoRR}, abs/2304.00723.

\bibitem[{Deriu et~al.(2021)Deriu, Rodrigo, Otegi, Echegoyen, Rosset, Agirre,
  and Cieliebak}]{DBLP:journals/air/DeriuROERAC21}
Jan Deriu, {\'{A}}lvaro Rodrigo, Arantxa Otegi, Guillermo Echegoyen, Sophie
  Rosset, Eneko Agirre, and Mark Cieliebak. 2021.
\newblock \href {https://doi.org/10.1007/s10462-020-09866-x} {Survey on
  evaluation methods for dialogue systems}.
\newblock \emph{Artif. Intell. Rev.}, 54(1):755--810.

\bibitem[{Fu et~al.(2023)Fu, Ng, Jiang, and
  Liu}]{DBLP:journals/corr/abs-2302-04166}
Jinlan Fu, See{-}Kiong Ng, Zhengbao Jiang, and Pengfei Liu. 2023.
\newblock \href {https://doi.org/10.48550/arXiv.2302.04166} {Gptscore: Evaluate
  as you desire}.
\newblock \emph{CoRR}, abs/2302.04166.

\bibitem[{Ghazarian et~al.(2019)Ghazarian, Wei, Galstyan, and
  Peng}]{ghazarian-etal-2019-better}
Sarik Ghazarian, Johnny Wei, Aram Galstyan, and Nanyun Peng. 2019.
\newblock \href {https://doi.org/10.18653/v1/W19-2310} {Better automatic
  evaluation of open-domain dialogue systems with contextualized embeddings}.
\newblock In \emph{Proceedings of the Workshop on Methods for Optimizing and
  Evaluating Neural Language Generation}, pages 82--89, Minneapolis, Minnesota.
  Association for Computational Linguistics.

\bibitem[{Ghazarian et~al.(2020)Ghazarian, Weischedel, Galstyan, and
  Peng}]{DBLP:conf/aaai/GhazarianWGP20}
Sarik Ghazarian, Ralph~M. Weischedel, Aram Galstyan, and Nanyun Peng. 2020.
\newblock \href {https://ojs.aaai.org/index.php/AAAI/article/view/6283}
  {Predictive engagement: An efficient metric for automatic evaluation of
  open-domain dialogue systems}.
\newblock In \emph{The Thirty-Fourth {AAAI} Conference on Artificial
  Intelligence, {AAAI} 2020, The Thirty-Second Innovative Applications of
  Artificial Intelligence Conference, {IAAI} 2020, The Tenth {AAAI} Symposium
  on Educational Advances in Artificial Intelligence, {EAAI} 2020, New York,
  NY, USA, February 7-12, 2020}, pages 7789--7796. {AAAI} Press.

\bibitem[{Gopalakrishnan et~al.(2019)Gopalakrishnan, Hedayatnia, Chen,
  Gottardi, Kwatra, Venkatesh, Gabriel, and
  Hakkani{-}T{\"{u}}r}]{DBLP:conf/interspeech/GopalakrishnanH19}
Karthik Gopalakrishnan, Behnam Hedayatnia, Qinglang Chen, Anna Gottardi,
  Sanjeev Kwatra, Anu Venkatesh, Raefer Gabriel, and Dilek Hakkani{-}T{\"{u}}r.
  2019.
\newblock \href {https://doi.org/10.21437/Interspeech.2019-3079} {Topical-chat:
  Towards knowledge-grounded open-domain conversations}.
\newblock In \emph{Interspeech 2019, 20th Annual Conference of the
  International Speech Communication Association, Graz, Austria, 15-19
  September 2019}, pages 1891--1895. {ISCA}.

\bibitem[{Hori and Hori(2017)}]{DBLP:journals/corr/HoriH17}
Chiori Hori and Takaaki Hori. 2017.
\newblock \href {http://arxiv.org/abs/1706.07440} {End-to-end conversation
  modeling track in {DSTC6}}.
\newblock \emph{CoRR}, abs/1706.07440.

\bibitem[{Huang et~al.(2020)Huang, Ye, Qin, Lin, and
  Liang}]{huang-etal-2020-grade}
Lishan Huang, Zheng Ye, Jinghui Qin, Liang Lin, and Xiaodan Liang. 2020.
\newblock \href {https://doi.org/10.18653/v1/2020.emnlp-main.742} {{GRADE}:
  Automatic graph-enhanced coherence metric for evaluating open-domain dialogue
  systems}.
\newblock In \emph{Proceedings of the 2020 Conference on Empirical Methods in
  Natural Language Processing (EMNLP)}, pages 9230--9240, Online. Association
  for Computational Linguistics.

\bibitem[{Kong{-}Vega et~al.(2018)Kong{-}Vega, Shen, Wang, and
  D'Haro}]{DBLP:conf/iwsds/Kong-VegaSWD18}
Naomi Kong{-}Vega, Mingxin Shen, Mo~Wang, and Luis~Fernando D'Haro. 2018.
\newblock \href {https://doi.org/10.1007/978-981-13-9443-0\_32} {Subjective
  annotation and evaluation of three different chatbots {WOCHAT:} shared task
  report}.
\newblock In \emph{9th International Workshop on Spoken Dialogue System
  Technology, {IWSDS} 2018, Singapore, April 18-20, 2018}, volume 579 of
  \emph{Lecture Notes in Electrical Engineering}, pages 371--378. Springer.

\bibitem[{Lee et~al.(2020)Lee, Lim, and
  Sedoc}]{DBLP:journals/corr/abs-2010-12741}
Seolhwa Lee, Heuiseok Lim, and Jo{\~{a}}o Sedoc. 2020.
\newblock \href {http://arxiv.org/abs/2010.12741} {An evaluation protocol for
  generative conversational systems}.
\newblock \emph{CoRR}, abs/2010.12741.

\bibitem[{Li et~al.(2021)Li, Zhang, Fei, Feng, and
  Zhou}]{li-etal-2021-conversations}
Zekang Li, Jinchao Zhang, Zhengcong Fei, Yang Feng, and Jie Zhou. 2021.
\newblock \href {https://doi.org/10.18653/v1/2021.acl-long.11} {Conversations
  are not flat: Modeling the dynamic information flow across dialogue
  utterances}.
\newblock In \emph{Proceedings of the 59th Annual Meeting of the Association
  for Computational Linguistics and the 11th International Joint Conference on
  Natural Language Processing (Volume 1: Long Papers)}, pages 128--138, Online.
  Association for Computational Linguistics.

\bibitem[{Lin(2004)}]{lin-2004-rouge}
Chin-Yew Lin. 2004.
\newblock \href {https://aclanthology.org/W04-1013} {{ROUGE}: A package for
  automatic evaluation of summaries}.
\newblock In \emph{Text Summarization Branches Out}, pages 74--81, Barcelona,
  Spain. Association for Computational Linguistics.

\bibitem[{Liu et~al.(2016)Liu, Lowe, Serban, Noseworthy, Charlin, and
  Pineau}]{liu-etal-2016-evaluate}
Chia-Wei Liu, Ryan Lowe, Iulian Serban, Mike Noseworthy, Laurent Charlin, and
  Joelle Pineau. 2016.
\newblock \href {https://doi.org/10.18653/v1/D16-1230} {How {NOT} to evaluate
  your dialogue system: An empirical study of unsupervised evaluation metrics
  for dialogue response generation}.
\newblock In \emph{Proceedings of the 2016 Conference on Empirical Methods in
  Natural Language Processing}, pages 2122--2132, Austin, Texas. Association
  for Computational Linguistics.

\bibitem[{Liu et~al.(2023)Liu, Iter, Xu, Wang, Xu, and
  Zhu}]{DBLP:journals/corr/abs-2303-16634}
Yang Liu, Dan Iter, Yichong Xu, Shuohang Wang, Ruochen Xu, and Chenguang Zhu.
  2023.
\newblock \href {https://doi.org/10.48550/arXiv.2303.16634} {G-eval: {NLG}
  evaluation using {GPT-4} with better human alignment}.
\newblock \emph{CoRR}, abs/2303.16634.

\bibitem[{Mehri and
  Eskenazi(2020{\natexlab{a}})}]{mehri-eskenazi-2020-unsupervised}
Shikib Mehri and Maxine Eskenazi. 2020{\natexlab{a}}.
\newblock \href {https://aclanthology.org/2020.sigdial-1.28} {Unsupervised
  evaluation of interactive dialog with {D}ialo{GPT}}.
\newblock In \emph{Proceedings of the 21th Annual Meeting of the Special
  Interest Group on Discourse and Dialogue}, pages 225--235, 1st virtual
  meeting. Association for Computational Linguistics.

\bibitem[{Mehri and Eskenazi(2020{\natexlab{b}})}]{mehri-eskenazi-2020-usr}
Shikib Mehri and Maxine Eskenazi. 2020{\natexlab{b}}.
\newblock \href {https://doi.org/10.18653/v1/2020.acl-main.64} {{USR}: An
  unsupervised and reference free evaluation metric for dialog generation}.
\newblock In \emph{Proceedings of the 58th Annual Meeting of the Association
  for Computational Linguistics}, pages 681--707, Online. Association for
  Computational Linguistics.

\bibitem[{Mehri et~al.(2022)Mehri, Feng, Gordon, Alavi, Traum, and
  Eskenazi}]{mehri-etal-2022-interactive}
Shikib Mehri, Yulan Feng, Carla Gordon, Seyed~Hossein Alavi, David Traum, and
  Maxine Eskenazi. 2022.
\newblock \href {https://aclanthology.org/2022.lrec-1.616} {Interactive
  evaluation of dialog track at {DSTC}9}.
\newblock In \emph{Proceedings of the Thirteenth Language Resources and
  Evaluation Conference}, pages 5731--5738, Marseille, France. European
  Language Resources Association.

\bibitem[{OpenAI(2023)}]{DBLP:journals/corr/abs-2303-08774}
OpenAI. 2023.
\newblock \href {https://doi.org/10.48550/arXiv.2303.08774} {{GPT-4} technical
  report}.
\newblock \emph{CoRR}, abs/2303.08774.

\bibitem[{Papineni et~al.(2002)Papineni, Roukos, Ward, and
  Zhu}]{papineni-etal-2002-bleu}
Kishore Papineni, Salim Roukos, Todd Ward, and Wei-Jing Zhu. 2002.
\newblock \href {https://doi.org/10.3115/1073083.1073135} {{B}leu: a method for
  automatic evaluation of machine translation}.
\newblock In \emph{Proceedings of the 40th Annual Meeting of the Association
  for Computational Linguistics}, pages 311--318, Philadelphia, Pennsylvania,
  USA. Association for Computational Linguistics.

\bibitem[{Phy et~al.(2020)Phy, Zhao, and Aizawa}]{phy-etal-2020-deconstruct}
Vitou Phy, Yang Zhao, and Akiko Aizawa. 2020.
\newblock \href {https://doi.org/10.18653/v1/2020.coling-main.368} {Deconstruct
  to reconstruct a configurable evaluation metric for open-domain dialogue
  systems}.
\newblock In \emph{Proceedings of the 28th International Conference on
  Computational Linguistics}, pages 4164--4178, Barcelona, Spain (Online).
  International Committee on Computational Linguistics.

\bibitem[{Rashkin et~al.(2019)Rashkin, Smith, Li, and
  Boureau}]{rashkin-etal-2019-towards}
Hannah Rashkin, Eric~Michael Smith, Margaret Li, and Y-Lan Boureau. 2019.
\newblock \href {https://doi.org/10.18653/v1/P19-1534} {Towards empathetic
  open-domain conversation models: A new benchmark and dataset}.
\newblock In \emph{Proceedings of the 57th Annual Meeting of the Association
  for Computational Linguistics}, pages 5370--5381, Florence, Italy.
  Association for Computational Linguistics.

\bibitem[{Sai et~al.(2020)Sai, Mohankumar, Arora, and
  Khapra}]{sai-etal-2020-improving}
Ananya~B. Sai, Akash~Kumar Mohankumar, Siddhartha Arora, and Mitesh~M. Khapra.
  2020.
\newblock \href {https://doi.org/10.1162/tacl_a_00347} {Improving dialog
  evaluation with a multi-reference adversarial dataset and large scale
  pretraining}.
\newblock \emph{Transactions of the Association for Computational Linguistics},
  8:810--827.

\bibitem[{Sedoc et~al.(2019)Sedoc, Ippolito, Kirubarajan, Thirani, Ungar, and
  Callison-Burch}]{sedoc-etal-2019-chateval}
Jo{\~a}o Sedoc, Daphne Ippolito, Arun Kirubarajan, Jai Thirani, Lyle Ungar, and
  Chris Callison-Burch. 2019.
\newblock \href {https://doi.org/10.18653/v1/N19-4011} {{C}hat{E}val: A tool
  for chatbot evaluation}.
\newblock In \emph{Proceedings of the 2019 Conference of the North {A}merican
  Chapter of the Association for Computational Linguistics (Demonstrations)},
  pages 60--65, Minneapolis, Minnesota. Association for Computational
  Linguistics.

\bibitem[{Smith et~al.(2022)Smith, Hsu, Qian, Roller, Boureau, and
  Weston}]{smith-etal-2022-human}
Eric Smith, Orion Hsu, Rebecca Qian, Stephen Roller, Y-Lan Boureau, and Jason
  Weston. 2022.
\newblock \href {https://doi.org/10.18653/v1/2022.nlp4convai-1.8} {Human
  evaluation of conversations is an open problem: comparing the sensitivity of
  various methods for evaluating dialogue agents}.
\newblock In \emph{Proceedings of the 4th Workshop on NLP for Conversational
  AI}, pages 77--97, Dublin, Ireland. Association for Computational
  Linguistics.

\bibitem[{Tao et~al.(2018)Tao, Mou, Zhao, and Yan}]{DBLP:conf/aaai/TaoMZY18}
Chongyang Tao, Lili Mou, Dongyan Zhao, and Rui Yan. 2018.
\newblock \href
  {https://www.aaai.org/ocs/index.php/AAAI/AAAI18/paper/view/16179} {{RUBER:}
  an unsupervised method for automatic evaluation of open-domain dialog
  systems}.
\newblock In \emph{Proceedings of the Thirty-Second {AAAI} Conference on
  Artificial Intelligence, (AAAI-18), the 30th innovative Applications of
  Artificial Intelligence (IAAI-18), and the 8th {AAAI} Symposium on
  Educational Advances in Artificial Intelligence (EAAI-18), New Orleans,
  Louisiana, USA, February 2-7, 2018}, pages 722--729. {AAAI} Press.

\bibitem[{Vinyals and Le(2015)}]{DBLP:journals/corr/VinyalsL15}
Oriol Vinyals and Quoc~V. Le. 2015.
\newblock \href {http://arxiv.org/abs/1506.05869} {A neural conversational
  model}.
\newblock \emph{CoRR}, abs/1506.05869.

\bibitem[{Wei et~al.(2022)Wei, Wang, Schuurmans, Bosma, Ichter, Xia, Chi, Le,
  and Zhou}]{DBLP:conf/nips/Wei0SBIXCLZ22}
Jason Wei, Xuezhi Wang, Dale Schuurmans, Maarten Bosma, Brian Ichter, Fei Xia,
  Ed~H. Chi, Quoc~V. Le, and Denny Zhou. 2022.
\newblock \href
  {http://papers.nips.cc/paper\_files/paper/2022/hash/9d5609613524ecf4f15af0f7b31abca4-Abstract-Conference.html}
  {Chain-of-thought prompting elicits reasoning in large language models}.
\newblock In \emph{NeurIPS}.

\bibitem[{Yeh et~al.(2021)Yeh, Eskenazi, and
  Mehri}]{yeh-etal-2021-comprehensive}
Yi-Ting Yeh, Maxine Eskenazi, and Shikib Mehri. 2021.
\newblock \href {https://doi.org/10.18653/v1/2021.eancs-1.3} {A comprehensive
  assessment of dialog evaluation metrics}.
\newblock In \emph{The First Workshop on Evaluations and Assessments of Neural
  Conversation Systems}, pages 15--33, Online. Association for Computational
  Linguistics.

\bibitem[{Zhang et~al.(2021{\natexlab{a}})Zhang, Chen, D{'}Haro, Zhang,
  Friedrichs, Lee, and Li}]{zhang-etal-2021-dynaeval}
Chen Zhang, Yiming Chen, Luis~Fernando D{'}Haro, Yan Zhang, Thomas Friedrichs,
  Grandee Lee, and Haizhou Li. 2021{\natexlab{a}}.
\newblock \href {https://doi.org/10.18653/v1/2021.acl-long.441} {{D}yna{E}val:
  Unifying turn and dialogue level evaluation}.
\newblock In \emph{Proceedings of the 59th Annual Meeting of the Association
  for Computational Linguistics and the 11th International Joint Conference on
  Natural Language Processing (Volume 1: Long Papers)}, pages 5676--5689,
  Online. Association for Computational Linguistics.

\bibitem[{Zhang et~al.(2020{\natexlab{a}})Zhang, D'Haro, Banchs, Friedrichs,
  and Li}]{DBLP:conf/iwsds/ZhangDBFL20}
Chen Zhang, Luis~Fernando D'Haro, Rafael~E. Banchs, Thomas Friedrichs, and
  Haizhou Li. 2020{\natexlab{a}}.
\newblock \href {https://doi.org/10.1007/978-981-15-8395-7\_5} {Deep {AM-FM:}
  toolkit for automatic dialogue evaluation}.
\newblock In \emph{Conversational Dialogue Systems for the Next Decade - 11th
  International Workshop on Spoken Dialogue Systems, {IWSDS} 2020, Madrid,
  Spain, 21-23 September, 2020}, volume 704 of \emph{Lecture Notes in
  Electrical Engineering}, pages 53--69. Springer.

\bibitem[{Zhang et~al.(2021{\natexlab{b}})Zhang, Sedoc, D'Haro, Banchs, and
  Rudnicky}]{DBLP:journals/corr/abs-2111-02110}
Chen Zhang, Jo{\~{a}}o Sedoc, Luis~Fernando D'Haro, Rafael~E. Banchs, and
  Alexander Rudnicky. 2021{\natexlab{b}}.
\newblock \href {http://arxiv.org/abs/2111.02110} {Automatic evaluation and
  moderation of open-domain dialogue systems}.
\newblock \emph{CoRR}, abs/2111.02110.

\bibitem[{Zhang et~al.(2022)Zhang, Hu, Yu, Wang, Han, Liu, and
  Yuan}]{DBLP:journals/corr/abs-2206-09403}
Pengfei Zhang, Xiaohui Hu, Kaidong Yu, Jian Wang, Song Han, Cao Liu, and
  Chunyang Yuan. 2022.
\newblock \href {https://doi.org/10.48550/arXiv.2206.09403} {{MME-CRS:}
  multi-metric evaluation based on correlation re-scaling for evaluating
  open-domain dialogue}.
\newblock \emph{CoRR}, abs/2206.09403.

\bibitem[{Zhang et~al.(2018)Zhang, Dinan, Urbanek, Szlam, Kiela, and
  Weston}]{zhang-etal-2018-personalizing}
Saizheng Zhang, Emily Dinan, Jack Urbanek, Arthur Szlam, Douwe Kiela, and Jason
  Weston. 2018.
\newblock \href {https://doi.org/10.18653/v1/P18-1205} {Personalizing dialogue
  agents: {I} have a dog, do you have pets too?}
\newblock In \emph{Proceedings of the 56th Annual Meeting of the Association
  for Computational Linguistics (Volume 1: Long Papers)}, pages 2204--2213,
  Melbourne, Australia. Association for Computational Linguistics.

\bibitem[{Zhang et~al.(2020{\natexlab{b}})Zhang, Kishore, Wu, Weinberger, and
  Artzi}]{DBLP:conf/iclr/ZhangKWWA20}
Tianyi Zhang, Varsha Kishore, Felix Wu, Kilian~Q. Weinberger, and Yoav Artzi.
  2020{\natexlab{b}}.
\newblock \href {https://openreview.net/forum?id=SkeHuCVFDr} {Bertscore:
  Evaluating text generation with {BERT}}.
\newblock In \emph{8th International Conference on Learning Representations,
  {ICLR} 2020, Addis Ababa, Ethiopia, April 26-30, 2020}. OpenReview.net.

\end{thebibliography}
\bibliographystyle{acl_natbib}

\appendix

\section{Prompt Templates}
\label{sec:appendix_prompt}

Below are the prompt templates used in our experiments with \textsc{LLM-Eval}. They provide examples of the natural language instructions used to define the evaluation task and desired criteria, as well as the format instructions that specify the structure and range of scores for each dimension.

\subsection{Evaluation Schema}

The evaluation schema used in \textsc{LLM-Eval} is a natural language instruction that defines the task and the desired evaluation criteria. It covers multiple dimensions of evaluation, such as content, grammar, relevance, and appropriateness. An example of the format instruction specifying the structure and range of scores for each dimension is as follows:

\begin{tcolorbox}[width=\columnwidth,colback=white]
\small
\begin{verbatim}
Human: The output should be formatted as a 
JSON instance that conforms to the JSON 
schema below.

As an example, for the schema {"properties": 
{"foo": {"title": "Foo", "description": "a 
list of strings", "type": "array", "items": 
{"type": "string"}}}, "required": ["foo"]}}
the object {"foo": ["bar", "baz"]} is a 
well-formatted instance of the schema. 
The object {"properties": {"foo": ["bar", 
"baz"]}} is not well-formatted.

Here is the output schema:
{"properties": {"content": {"title": 
"Content", "description": "content score 
in the range of 0 to 100", "type": 
"integer"}, "grammar": {"title": "Grammar", 
"description": "grammar score in the range 
of 0 to 100", "type": "integer"}, "relevance": 
{"title": "Relevance", "description": 
"relevance score in the range of 0 to 100", 
"type": "integer"}, "appropriateness": 
{"title": "Appropriateness", "description": 
"appropriateness  score in the range of 0 to 
100", "type": "integer"}}, "required": 
["content", "grammar", "relevance", 
"appropriateness"]}
\end{verbatim}
\end{tcolorbox}

\subsection{Reference-based Turn-level Evaluation}

For reference-based turn-level evaluation, the single prompt is designed to include the necessary dialogue context, the reference, and the target response that needs to be evaluated, along with the evaluation schema. An example prompt template for evaluating a dialogue response with a human reference is:

\begin{tcolorbox}[width=\columnwidth,colback=white]
\begin{verbatim}
{evaluation_schema}

Score the following dialogue response 
generated on a continuous scale from 
{score_min} to {score_max}.

Context: {context}
Reference: {reference}
Dialogue response: {response}
\end{verbatim}
\end{tcolorbox}

\subsection{Reference-free Turn-level Evaluation}

For reference-free turn-level evaluation, the single prompt includes the dialogue context and the target response that needs to be evaluated, without requiring a human reference. The evaluation schema is also included in the prompt. An example prompt template for evaluating a dialogue response without a human reference is:

\begin{tcolorbox}[width=\columnwidth,colback=white]
\begin{verbatim}
{evaluation_schema}

Score the following dialogue response 
generated on a continuous scale from 
{score_min} to {score_max}.

Context: {context}
Dialogue response: {response}
\end{verbatim}
\end{tcolorbox}

\subsection{Dialogue-level Evaluation}

For dialogue-level evaluation, the single prompt is designed to cover the entire dialogue instead of individual turns. The evaluation schema is also included in the prompt. An example prompt template for evaluating a dialogue is:

\begin{tcolorbox}[width=\columnwidth,colback=white]
\begin{verbatim}
{evaluation_schema}

Score the following dialogue generated 
on a continuous scale from {score_min} 
to {score_max}.

Dialogue: {dialog}
\end{verbatim}
\end{tcolorbox}

\end{document}